\pdfoutput=1

\documentclass[11pt]{article}

\usepackage{EMNLP2022}
\usepackage{times}
\usepackage{latexsym}
\usepackage{setspace}
\usepackage{color}
\usepackage{array}
\usepackage[normalem]{ulem}

\newcommand{\blue}[1]{\textcolor{blue}{#1}}

\newcolumntype{"}{@{\hskip\tabcolsep\vrule width 1pt\hskip\tabcolsep}}
\usepackage[T1]{fontenc}

\usepackage[utf8]{inputenc}
\usepackage{graphicx}
\usepackage{microtype}
\usepackage{booktabs}
\usepackage{amsmath,amsthm,amsfonts,amssymb,bm}
\usepackage{mathrsfs}
\usepackage{amsmath}
\usepackage{algorithm}  
\usepackage{algorithmicx}  
\usepackage{algpseudocode}  
\usepackage{booktabs}
\usepackage{enumitem}
\usepackage{amsthm}
\usepackage{amssymb}

\theoremstyle{definition}
\usepackage{multirow}
\usepackage{verbatim}
\usepackage{etoolbox}
\usepackage{hyperref}
\usepackage{subfigure}
\makeatletter
\def\UrlAlphabet{%
      \do\a\do\b\do\c\do\d\do\e\do\f\do\g\do\h\do\i\do\j%
      \do\k\do\l\do\m\do\n\do\o\do\p\do\q\do\r\do\s\do\t%
      \do\u\do\v\do\w\do\x\do\y\do\z\do\A\do\B\do\C\do\D%
      \do\E\do\F\do\G\do\H\do\I\do\J\do\K\do\L\do\M\do\N%
      \do\O\do\P\do\Q\do\R\do\S\do\T\do\U\do\V\do\W\do\X%
      \do\Y\do\Z}
\def\UrlDigits{\do\1\do\2\do\3\do\4\do\5\do\6\do\7\do\8\do\9\do\0}
\g@addto@macro{\UrlBreaks}{\UrlOrds}
\g@addto@macro{\UrlBreaks}{\UrlAlphabet}
\g@addto@macro{\UrlBreaks}{\UrlDigits}
\usepackage[utf8]{inputenc}
\usepackage[T1]{fontenc}

\usepackage{titlesec}
\titlespacing*{\paragraph}{0pt}{4pt}{4pt}

\usepackage{cleveref}
\crefname{section}{§}{§§}
\Crefname{section}{§}{§§}

\usepackage{microtype}

\DeclareMathOperator*{\argtopk}{arg\,topk} 
\newcommand{\thickhline}{
    \noalign {\ifnum 0=`}\fi \hrule height 1pt
    \futurelet \reserved@a \@xhline
}
\usepackage{tikz}
\newcommand*{\circled}[1]{\lower.7ex\hbox{\tikz\draw (0pt, 0pt)%
    circle (.5em) node {\makebox[1em][c]{\small #1}};}}
\setitemize[1]{leftmargin=10pt,itemsep=0.5pt,partopsep=0pt,parsep=0.5pt,topsep=0pt}

\title{Improving Multi-turn Emotional Support Dialogue Generation with Lookahead Strategy Planning}

\author{
Yi Cheng$^{1}$\thanks{\ \ Equal contribution.}\ \ , Wenge Liu$^{2}$\footnotemark[1]\ \ , Wenjie Li$^{1}$\thanks{\ \ Corresponding author.}\ \ , Jiashuo Wang$^{1}$, \\\textbf{Ruihui Zhao$^{3}$, Bang Liu$^{4}, $Xiaodan Liang$^{5}$, Yefeng Zheng$^{3}$}\\
$^1$Hong Kong Polytechnic University \quad
$^2$Baidu Inc., Beijing, China \quad
$^3$Tencent Jarvis Lab\\
$^4$RALI \& Mila, Université de Montréal \quad
$^5$Sun Yat-sen University\\
\tt \{csycheng,cswjli,jessiejs.wang\}@comp.polyu.edu.hk,\\ 
\tt \{kzllwg,xdliang328\}@gmail.com, bang.liu@umontreal.ca,  \\ 
\tt \{zacharyzhao,yefengzheng\}@tencent.com \\
}

\begin{document}
\maketitle

\begin{abstract}
Providing Emotional Support (ES) to soothe people in emotional distress is an essential capability in social interactions. 
Most existing research on building ES conversation systems only considers single-turn interactions with users, which is over-simplified. 
In comparison, multi-turn ES conversation systems can provide ES more effectively, but face several new technical challenges, including: 
i) how to conduct support strategy planning that could lead to the best supporting effects; 
ii) how to dynamically model the user's state. 
In this paper, we propose a novel system named MultiESC to address these issues. 
For strategy planning, drawing inspiration from the A* search algorithm, we propose lookahead heuristics to estimate the future user feedback after using particular strategies, which helps to select strategies that can lead to the best long-term effects. For user state modeling, {MultiESC} focuses on capturing users' subtle emotional expressions and understanding their emotion causes. 
Extensive experiments show that {MultiESC} significantly outperforms competitive baselines in both strategy planning and dialogue generation. Our codes are available at \url{https://github.com/lwgkzl/MultiESC}. 
 
\end{abstract}
\section{Introduction}

\begin{figure}[t]
    \centering
    \vspace{3mm}
    \includegraphics[width=\linewidth]{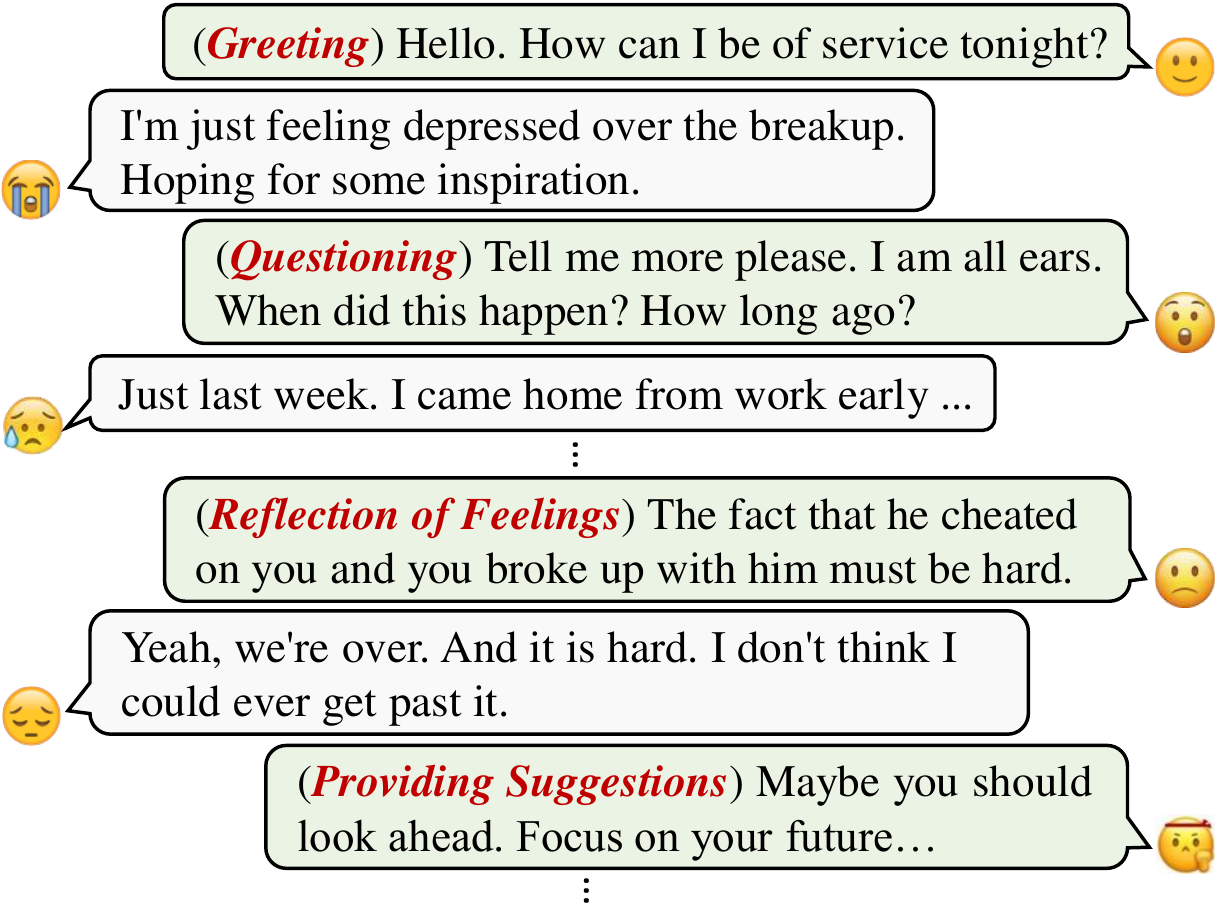}
    \vspace{-3mm}
    \caption{An example of an emotional support conversation between the support-seeker (left) and the supporter (right). The support strategies adopted by the supporter are presented in red italics before the utterances. }
     \vspace{-2mm}
    \label{fig:intro}
\end{figure}

Almost every human has experienced emotional distress, even if not suffering from any mental disorders. Frequently, people deal with the distress by seeking Emotional Support (ES) from social interactions \cite{langford1997social,greene2003handbook}. 
Nevertheless, ES from family and friends is not always available \cite{webber2018disaster}. 
With the potential of providing more people with in-time support, developing Emotional Support Conversation (ESC) systems has attracted much attention. 
However, since early ES datasets are constructed by crawling post-response pairs from online forums, they only contain single-turn conversations  \cite{medeiros2018using,sharma2020computational}. Thus, most of the existing research on ESC also only considers single-turn interactions with the user \cite{medeiros2018using,sharma2020computational,sharma2021towards}, which is over-simplified and has limited support effects. 
It was not until recently that \citet{liu2021towards} released the first large-scale multi-turn ES dataset, \textsc{ESConv}. They also designed an ESC framework, 
suggesting the conversation procedures and support strategies for multi-turn ESC. 

Compared to the single-turn scenario, developing multi-turn ESC systems faces several new challenges. 
One significant challenge is \emph{support strategy planning}. As pointed out in the psychological literature, particular procedures and strategies are indispensable for effective emotional support \cite{greene2003handbook, hill2009helping}. 
As in Fig. \ref{fig:intro}, the supporter strategically soothes the support-seeker by first caringly inquiring about the situation, then resonating with the seeker's feelings, and finally providing suggestions to evoke positive emotions. 



Notably, strategy planning in ESC should be conducted on a long planning horizon. 
That is, instead of merely considering the dialogue history or foreseeing the immediate effect after using the strategy, the system should further \emph{look ahead}, to consider how much the adopted strategy would contribute to reducing the user's emotional distress at a long run. 
Though some strategies may not directly provide comfort, they are still essential for reaching the long-term dialogue goal, such as greetings at the beginning of the conversation and inquiring about the user’s experiences. 

Another challenge for multi-turn ESC is how to \emph{dynamically model the user's state} during the conversation. Prior works on emotion-related dialogue tasks mainly detect the user's coarse-grained emotion type  to enhance dialogue generation \cite{moel,majumder2020mime,li2020empdg}. However, such practice is not completely appropriate for ESC. 
The reason is that the user’s emotion in ESC almost stays the same type, such as being sad, throughout the conversation. Instead, it often changes subtly in terms of emotion intensity. 
Besides, effective ES requires more than only identifying the user's emotion. A thorough understanding of the user's situation is also essential. 

In this paper, we propose a multi-turn ESC system {MultiESC} to address the above issues.
For \emph{strategy planning}, we draw inspiration from the $\text{A}^*$ search algorithm \cite{hart1968formal,pearl1985heuristics} and its recent application in constrained text generation \cite{lu2021neurologic}, which addressed the challenge of planning ahead by incorporating heuristic estimation of future cost. 
In {MultiESC}, we develop lookahead heuristics to estimate the expectation of the future user feedback to help select the strategy that can lead to the best long-term effect. Concretely, we implement a strategy sequence generator to produce the probability of the future strategy sequences, and a user feedback predictor to predict the feedback after applying the sequence of strategies. 
For \emph{user state modeling}, {MultiESC} captures the user's subtle emotion expressed in the context by incorporating external knowledge from the NRC VAD lexicon \cite{mohammad-2018-obtaining}. Moreover, it identifies the user's emotion causes (i.e., the experiences that caused the depressed emotion) to more thoroughly understand the user's situation.

In summary, our contributions are as follows:
\begin{itemize}
    \item We propose a multi-turn ESC system, MultiESC, which conducts support strategy planning with foresight of the user feedback and dynamically tracks the user's state by capturing the subtle emotional expressions and the emotion causes. 
    \item It is a pioneer work that adopts $\text{A}^*$-like lookahead heuristics to achieve dialogue strategy selection on a long planning horizon.
    \item Experiments show that MultiESC significantly outperforms a set of state-of-the-art models in generation quality and strategy planning, demonstrating the effectiveness of our proposed method. 
\end{itemize}

\section{Related Work}
\paragraph{Emotional Support Conversation Systems.}
Since early ES datasets were mainly composed of single-turn conversations  \cite{medeiros2018using,sharma2020computational}, most research on developing ESC systems only considered the simplified scenario of single-turn interactions with the user \cite{sharma2021towards,hosseini2021takes}. 
The few works that developed multi-turn ES chatbots rely on predefined templates and handcrafted rules \cite{zwaan2012conversation}, which suffer from limited generality. 
It was not until last year that \citet{liu2021towards} released the first multi-turn ESC dataset  \textsc{ESConv}. 
Following \citet{liu2021towards}, \citet{peng2022control} and \citet{tu2022misc} recently explored data-driven multi-turn ESC systems. 
\citet{peng2022control} proposed a hierarchical graph network to capture both the global context and the local user intention. They did not consider strategy planning, which is critical in multi-turn ESC. 
\citet{tu2022misc} proposed to enhance context encoding with commonsense knowledge and use the predicted strategy distribution to guide response generation. Nevertheless, their method of strategy prediction, directly implemented with a vanilla Transformer encoder, was relatively preliminary and did not consider any user-feedback-oriented planning as we do. 

\paragraph{Empathetic Response Generation.}
Empathetic Response Generation (ERG) \cite{rashkin2019towards} is a research area closely related to ESC, as being empathetic is a crucial ability for providing emotional support \cite{greene2003handbook,perez-rosas-etal-2017-understanding}. However, ERG does not has the explicit goal of proactively soothing the user's negative emotion. 
Instead, it only reactively generates responses that are consistent with the user's emotion \cite{moel,majumder2020mime,li2020empdg,zheng2021comae,wang2021empathetic}. 

\begin{figure}[t]
    \centering
    \includegraphics[width=\linewidth]{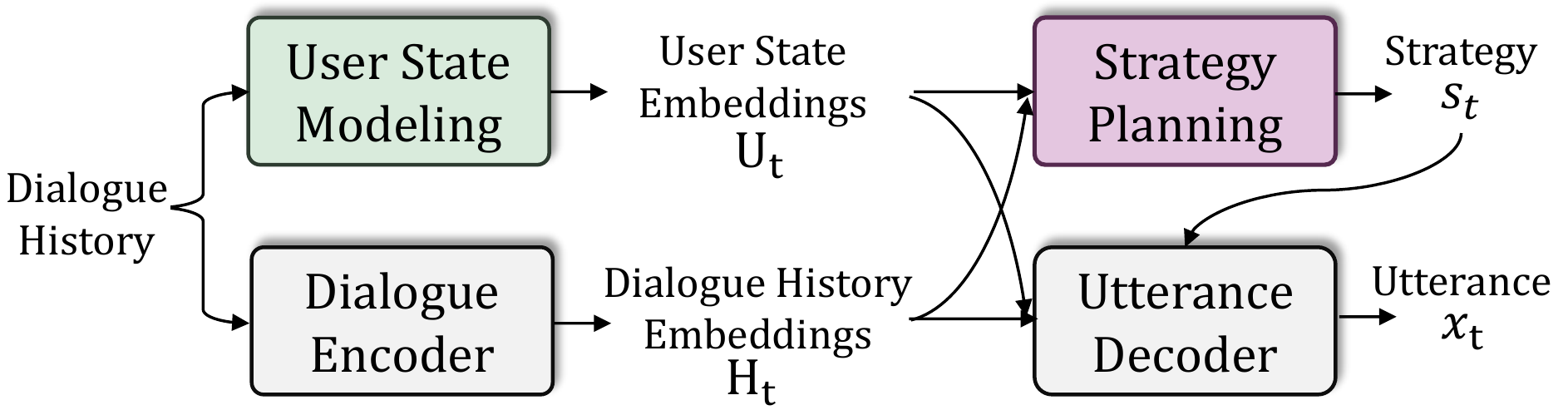}
    \caption{The overall framework of {MultiESC}. Details about the user state modeling and the strategy planning modules are shown in Fig. \ref{fig:user_state_modeling} and Fig.  \ref{fig:strategy_planning}, respectively.}
    \label{fig:overall_architecture}
\end{figure}

\section{Preliminaries}\label{sec:preliminaries}
\paragraph{ESConv.}
Our research is conducted on \textsc{ESConv}. It is a long conversation dataset, with an average of 29.8 utterances in each dialogue. 
It also includes rich annotations, such as the \emph{strategies} adopted by the supporter and the \emph{user feedback} scores. 
There are overall eight types of strategies (e.g., \emph{question}, \emph{reflection of feelings} and  \emph{self-disclosure}). 
The user feedback score indicates how much the user's emotional distress is reduced during the conversation. They are marked by the support-seekers on a Likert scale with five levels after every two turns. More data statistics are provided in the appendix. 

\paragraph{NRC VAD Lexicon.} The NRC VAD lexicon includes the Valence-Arousal-Dominance (VAD) scores of 20,000 English words. 
The VAD score of a word measures its underlying emotion in three dimensions: valence (pleased-displeased), 
arousal (excited-calm), 
and dominance (dominant-submissive). 
For example, the VAD scores of  ``loneliness'' and ``abandon'' are  (0.15, 0.18, 0.22) and (0.05, 0.52, 0.25), respectively.  
The VAD model captures a wide range of emotions and allows different emotions to be comparable. 


\paragraph{Problem Formulation of ESC.} Denote the utterances from the system and the user at the $i$-th round of the conversation are respectively $(x_i, y_i)$,\footnote{We suppose that ESCs are always initiated by the system (or the supporter).} while the user's state is $u_i$ ($i$=1, 2, ..., $n_{\text{R}}$). 
Suppose the set of all support strategies is $\mathcal{S}$. 
At the $t$-th turn, given the dialogue history  $\mathcal{H}_{t}$=$\{(x_i, y_i)\}_{i=1}^{t-1}$,  
the system tracks the user states $\mathcal{U}_t$=$\{u_1, u_2, ..., u_{t-1}\}$ from $\mathcal{H}_t$ and generates the next utterance $x_{t}$, using an appropriate support strategy $\hat{\mathbf{s}}_t\in \mathcal{S}$.

\section{Methodology}

As shown in Fig. \ref{fig:overall_architecture}, our proposed system  {MultiESC} consists of four modules. 
The dialogue encoder first converts  the dialogue history $\mathcal{H}_{t}$ into the embeddings $\mathbf{H}_t$. At the same time,  the user state modeling module extracts the user state information, producing the embeddings $\mathbf{U}_t$.
Then, given $\mathbf{H}_t$ and $\mathbf{U}_t$, the strategy planning module selects the strategy $s_t$. 
Finally, the utterance decoder generates the utterance $x_t$, adopting the strategy $s_{t}$. 

\subsection{Dialogue Encoder}
The dialogue encoder module is implemented with a Transformer encoder \cite{vaswani2017attention}. We concatenate the utterances in $\mathcal{H}_{t}$ and keep the last $N$ tokens of the concatenation as its input sequence. 
Given the input, it produces the dialogue history embeddings $\mathbf{H}_t$  $\in$ $\mathbb{R}^{N \times d_{\text{emb}}}$. 

\begin{figure}[t]
    \centering
    \includegraphics[width=\linewidth]{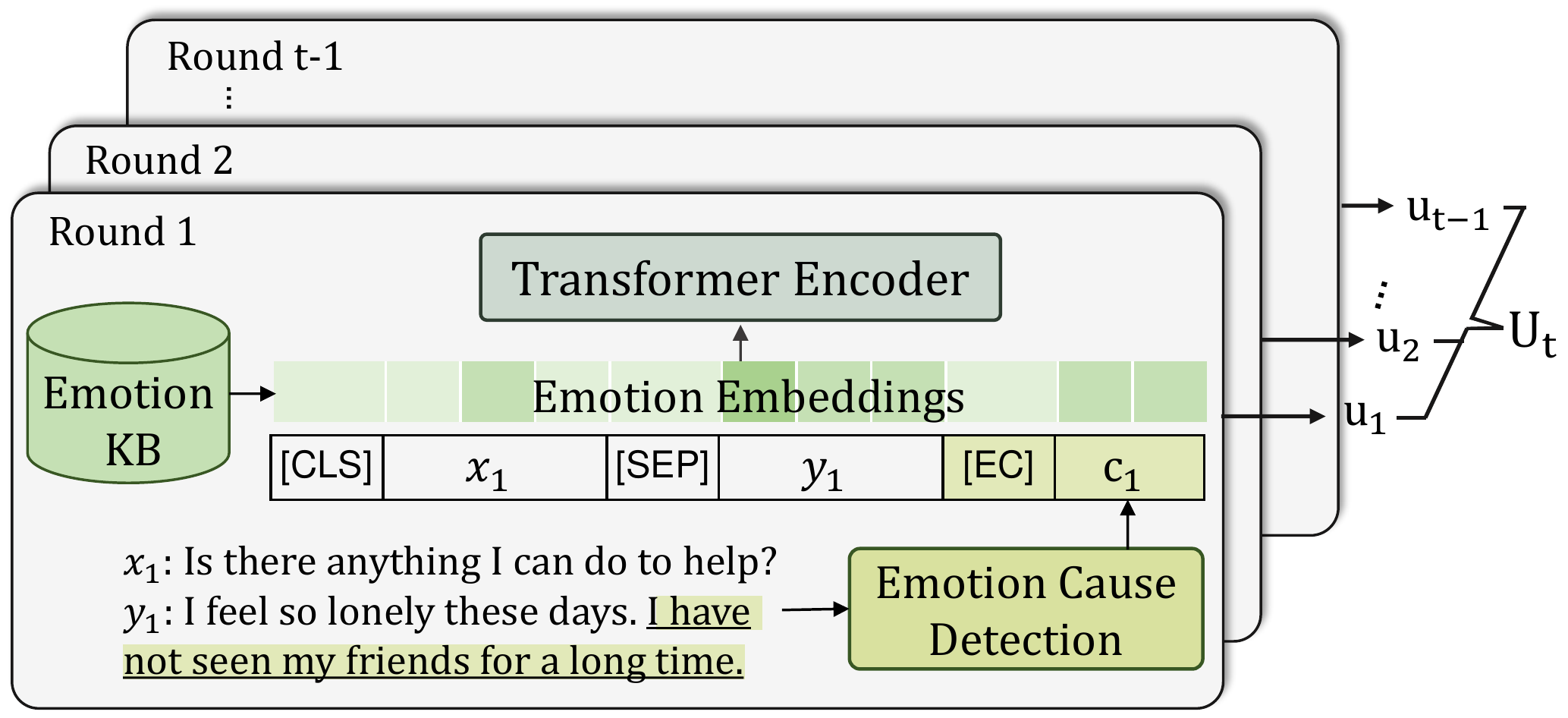}
    \caption{The architecture of the user state modeling module in {MultiESC}. }
    \label{fig:user_state_modeling}
\end{figure}

\begin{figure*}[t]
    \centering
    \includegraphics[width=0.95\linewidth]{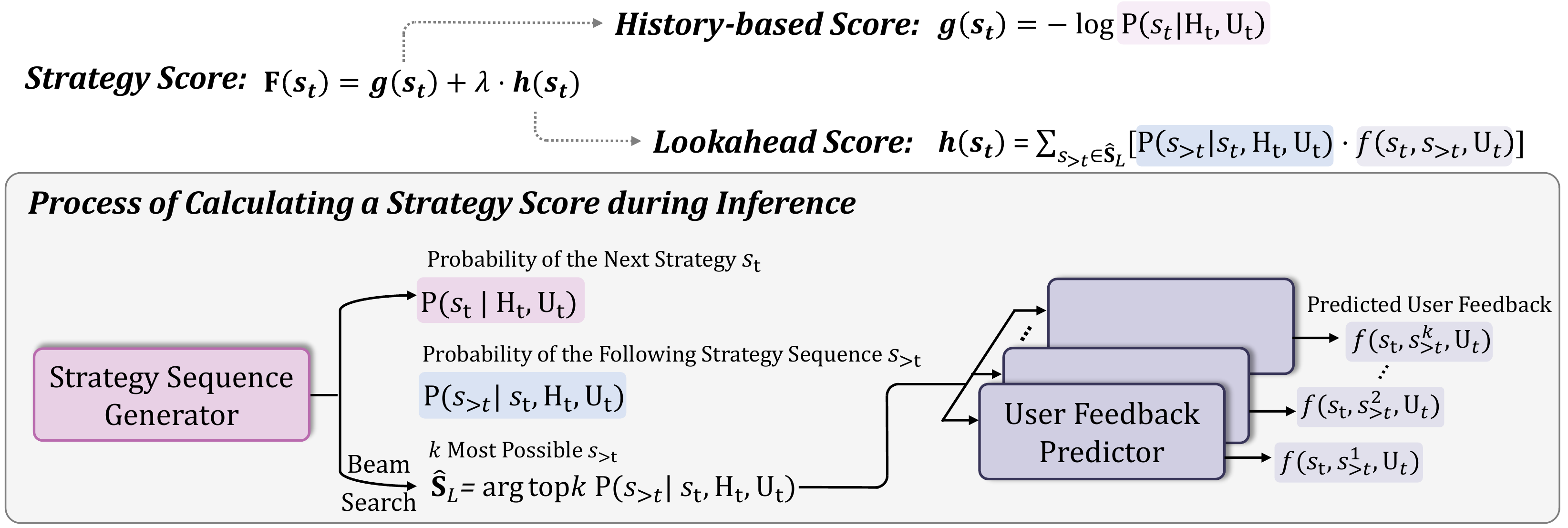}
    \caption{The process of calculating the strategy score, using a strategy sequence generator and a user feedback predictor. At each turn, our model selects the next strategy that maximizes the score of $F(s_t)$. }
    \label{fig:strategy_planning}
\end{figure*}

\subsection{User State Modeling}

Fig. \ref{fig:user_state_modeling} illustrates the workflow of user state modeling. 
To identify the user's state at the $i$-th round of the conversation, we first extract the emotion cause mentioned at this round, denoted as  $c_i$, with an off-the-shelf detector\footnote{\url{https://github.com/declare-lab/RECCON}} trained on a large-scale emotion cause detection dataset \cite{poria2021recognizing}. 
For example, in Fig. \ref{fig:user_state_modeling}, $c_1$=`` \emph{I have not seen my friends for a long time}''. 
Then, we concatenate the dialogue content $x_i$, $y_i$ and the emotion cause $c_i$ with special separator tokens to form the input of a Transformer encoder. Here, the system's utterance $x_i$ is also considered because it often provides necessary context for understanding the user's state. 
The input sequence is represented as the positional sum of emotion embeddings, word embeddings, and positional embeddings. 

The emotion embeddings are used to fuse the emotion information. They are obtained as follows.  
We train multiple emotion vectors $\{\mathbf{e}_1, \mathbf{e}_2, ..., \mathbf{e}_{n_{\text{emo}}}\}$ to represent the underlying emotions of different words. 
Concretely, we split the VAD space into multiple subspaces by dividing the valence and the arousal dimensions, respectively, into $n_{\text{V}}$ and $n_{\text{A}}$ intervals of equal length. 
Each emotional subspace is represented as one emotion vector $e_j$.\footnote{The dominance dimension is not considered here as it is less relevant for capturing emotion intensity \cite{zhong2019knowledge,li2020knowledge}.} 
To construct the emotion embeddings, we retrieve the VAD score of each input token from the NRC VAD lexicon to identify which emotional subspace it belongs, and then we represent it as the corresponding emotion vector.  
For those tokens without VAD annotation, we use a special emotion vector to represent them. 

Finally, the encoded hidden vector $\mathbf{u}_i$  corresponding to the $\texttt{[CLS]}$ token is used to represent the user state at the $i$-th round. The user state embeddings $\mathbf{U}_t$ is the concatenation of all the user state embeddings before the $t$-th round, that is, $\mathbf{U}_t$=$[\mathbf{u}_1; \mathbf{u}_2;...;\mathbf{u}_{t-1}]$.


\subsection{Strategy Planning with Lookahead Heuristics}
We develop a strategy score function to evaluate whether to adopt a particular strategy (e.g. \emph{question} or \emph{self-disclosure}) by comprehensively considering the dialogue history and the potential user feedback. 
Formally, at the $t$-th round, MultiESC adopts the strategy $\hat{\mathbf{s}}_t$ that maximizes the score function: 
\begin{equation}
    \hat{\mathbf{s}}_t=\mathop{\arg\max}_{{s}_t\in \mathcal{S}}F({s}_t),
\end{equation}
where $F(\cdot)$ is the strategy score function. 

In the following, we will first introduce the strategy score function and then explain how MultiESC calculates the strategy scores with two components: a strategy sequence generator and a user feedback predictor, as presented in Fig. \ref{fig:strategy_planning}. 
Finally, we will describe the architectures of the two components. 

\paragraph{Strategy Score Function.} \label{sec: strategy score function}
Our method draws inspiration from the classical search algorithm, $\text{A}^{*}$ search \cite{hart1968formal}, which conducts lookahead planning in a heuristic way. 
At each step, it searches the highest-scoring path by selecting an action that maximizes the sum of the score so far and a heuristic estimation of the future score. 
Similarly, we define our strategy score function as:
\begin{equation}
    F(s_{t}) = g(s_{t})+\lambda\cdot h(s_{t}),\label{eq:F(s_{t})}
\end{equation}
where $g(s_{t})$ is a \emph{history-based score};  $h(s_{t})$ is a \emph{lookahead score} that heuristically estimates the future user feedback; $\lambda$ is a hyper-parameter that balances the weights of the two terms. 

The {history-based score} $g(s_{t})$ computes the conditional probability distribution of the next strategy purely based on the dialogue history and the previous user states. Formally, it is defined as: 
\begin{equation}
    g(s_{t})=-\log \text{Pr}(s_{t}|\mathbf{H}_{t}, \mathbf{U}_{t}).\label{eq:g(s_{t})}
\end{equation}
Previous research on dialogue strategy prediction generally followed this history-based scheme \cite{zhou2019augmenting, joshi2021dialograph, dutt2021resper}, though they may vary in their methods of obtaining the representations of $\mathbf{H}_{t}$ and $\mathbf{U}_{t}$. However, such practice overlooks the strategy's future effects and how much it could help in achieving the long-term dialogue goal. In our work, we incorporate the lookahead score to alleviate this issue. 

The {lookahead score} $h(s_{t})$ heuristically estimates \emph{the mathematical expectation of the future user feedback score}\footnote{The \emph{user feedback score} indicates how much the user's emotional distress is reduced (see \Cref{sec:preliminaries}).} after adopting the strategy $s_t$ at the $t$-th round. 
Ideally, to select the strategy that could lead to the best final result, we want to estimate the user feedback score at the end of the conversation, that is:
\begin{equation}
\resizebox{\hsize}{!}{$
    \begin{aligned}
        h(s_{t}) &= \text{E}[f(s_{t},s_{{>t}}, \mathbf{U}_{t})|s_{t},\mathbf{H}_t,\mathbf{U}_t]\\
        &=\sum_{s_{_{>t}}\in \mathcal{S}^{\omega}}\text{Pr}(s_{{>t}}|s_{t},\mathbf{H}_t,\mathbf{U}_t)\cdot f(s_{t},s_{{>t}},\mathbf{U}_t). 
    \end{aligned}\label{eq:perfect_h(s_t)}$}
\end{equation}
E($\cdot$) represents the mathematical expectation;   $s_{{>t}}$ is the future strategy sequence to be used after the $t$-th round till the end of the conversation; $\mathcal{S}^{\omega}$ is the set of all possible strategy sequences; 
$f(s_{t},s_{>t}, \mathbf{U}_t)$ denotes the user feedback score after successively applying $s_{t}$ and $s_{{>t}}$ to comfort a user whose previous states are $\mathbf{U}_t$. 

However, Eq. \ref{eq:perfect_h(s_t)} is hard to calculate, because the space of $\mathcal{S}^{\omega}$ is too large and it is difficult to estimate the user feedback $f(\cdot)$ after too many turns (i.e. if the strategy sequence $s_{>t}$ is too long). 
Thus, we approximate Eq. \ref{eq:perfect_h(s_t)} as follows.
First, we only look ahead for the limited $L$ turns. We estimate the expectation of the user feedback score after $L$ turns instead of at the end of the conversation. Then, to further narrow the space of $\mathcal{S}^{\omega}$, we only consider the $k$ most possible future strategy sequences. 
Formally, Eq. \ref{eq:perfect_h(s_t)} is approximated as: 
\begin{equation}
\resizebox{\hsize}{!}{$
    \begin{aligned}
        h(s_{t}) &= \sum_{s_{_{>t}}\in \mathbf{\hat{\mathcal{S}}}_{L}}\text{Pr}(s_{{>t}}|s_{t},\mathbf{H}_t,\mathbf{U}_{t})\cdot f(s_{t}, s_{>t}, \mathbf{U}_t), \\
        \mathbf{\hat{\mathcal{S}}}_{{L}} &= \argtopk_{s_{{>t}}\in \mathbf{S}_{{L}}}\text{Pr}(s_{{>t}}|s_{t},\mathbf{H}_t,\mathbf{U}_t),
    \end{aligned}$}\label{eq:h(s_{t})}
\end{equation}
where $\mathcal{S}_L$ is the set of the strategy sequences whose lengths are less than $L$. 

\paragraph{Strategy Score Calculation in MultiESC.}
\label{sec:strategyMultiESC}
MultiESC calculates the strategy scores with a Strategy Sequence Generator (SSG) and a User Feedback Predictor (UFP). 
The function of SSG is to sequentially predict $s_{\ge t}$  based on $\mathbf{H}_{t}$ and $\mathbf{U}_{t}$, where $s_{\ge t}$ is the strategy sequence that will be used in the following $L$ rounds ($s_{\ge t}$=$[s_t; s_{>t}]$). 
At the $l$-th timestep, it outputs the predicted strategy distribution: 
\begin{equation}
\text{Pr}(s_{{t+l}}|s_{{t:t+l}},\textbf{H}_t,\textbf{U}_t),
\end{equation}
where $l$=$1, 2, ..., L$ and $s_{t:t+l}$ denotes the already-generated strategy sequence before the $l$-th timestep. The function of UFP is to estimate the user feedback score $f(s_{\ge t}, \mathbf{U}_t)$. 

As shown in Fig. \ref{fig:strategy_planning}, to calculate the strategy score of a particular strategy $s_t$, we first use SSG to derive the history-based score $g(s_{t})$ from its predicted strategy distribution at the first timestep. 
Next, we use SSG to find the set of the $k$ most possible future strategy sequences $\mathbf{\hat{S}}_{{L}}$ through beam search. For each strategy sequence $s_{>t}$ in $\mathbf{\hat{S}}_{{L}}$, we obtain its probability by: 
\begin{equation}
    \resizebox{\hsize}{!}{$\text{Pr}(s_{>t}|s_{t},\mathbf{H}_{t},\mathbf{U}_{t})=\prod_{l=2}^{L} \text{Pr}(s_{{t+l}}|s_{{t:t+l}},\mathbf{H}_{t},\mathbf{U}_{t}). $}\nonumber
\end{equation}
We then leverage UFP to estimate the user feedback score after successively applying $s_{t}$ and  $s_{>t}$. 
Combining the predicted probabilities of the strategy sequences and the estimated user feedback scores, we obtain the lookahead score $h(s_{t})$ as in Eq. \ref{eq:h(s_{t})}. 
Finally, given $g(s_{t})$ and $h(s_{t})$, the overall strategy score is obtained as in Eq. \ref{eq:F(s_{t})}.

\paragraph{Strategy Sequence Generator.}\label{sec:SSG}
SSG is developed upon the architecture of the Transformer decoder. 
Its only difference from the original Transformer decoder is that it adopts the multi-source attention mechanism to selectively attend to the dialogue history $\mathbf{H}_{t}$ and the user state information $\mathbf{U}_{t}$. 
Specifically, the strategy sequence $s_{\ge t}$ is first fed to a masked multi-head attention layer, 
producing the contextualized strategy sequence representations $\mathbf{P}_t$. 
Then, $\mathbf{P}_t$ interacts with  $\mathbf{H}_t$ and  $\mathbf{U}_t$ respectively through cross attention layers as:
\begin{align}
    \hat{\mathbf{H}}_t &= \texttt{MH-ATT}(L(\mathbf{P}_t), L(\mathbf{H}_t), L(\mathbf{H}_t)),\nonumber\\
    \hat{\mathbf{U}}_t &= \texttt{MH-ATT}(L(\mathbf{P}_t), L(\mathbf{U}_t), L(\mathbf{U}_t)),\nonumber
\end{align}
where $\texttt{MH-ATT}(\cdot)$ represents the multi-head self-attention mechanism. 
An information fusion layer is utilized to combine them:
\begin{align}
    \mu &= \text{ReLU}(\mathbf{W}_{\mu}[\hat{\mathbf{H}}_t; \hat{\mathbf{U}}_t]+\mathbf{b}_{\mu}), \nonumber\\
    \hat{\mathbf{P}}_t  &= \mu \cdot \hat{\mathbf{H}}_t + (1-\mu) \cdot  \hat{\mathbf{U}}_t, \nonumber
\end{align}
where $\mathbf{W}_{\mu}$ and $\mathbf{b}_{\mu}$ are trainable parameters. 
Next, $\hat{\mathbf{P}}_t$ is fed to a connected feed-forward network with residual connections around the sub-layers. We denote the hidden states produced here as  $\widetilde{\mathbf{P}}_t$.   
Finally, the strategy distribution at the $l$-th timestep is predicted as: 
\begin{equation}
    \text{Pr}(s_{{t+l}}|s_{{t:t+l}}, \mathbf{H}_t, \mathbf{U}_t)=\text{softmax}(\mathbf{W}_s\widetilde{\mathbf{P}}_t+\mathbf{b}_s),\nonumber
\end{equation}
where $\mathbf{W}_{s}$ and $\mathbf{b}_{s}$ are trainable parameters.
To train the SSG, we use the negative log-likelihood of the ground-truth strategy $s_{{t+l}}^{*}$ as its loss function. 

\begin{table*}[t]
\small
\centering
	\begin{center}\scalebox{1.0}{
		\begin{tabular}{ l  rccccccc } \specialrule{1.0pt}{1pt}{1pt}
        {\textbf{Model}} & {\textbf{PPL}$\downarrow$}& {\textbf{B-1}$\uparrow$} & {\textbf{B-2}$\uparrow$} & {\textbf{B-3}$\uparrow$} & {\textbf{B-4}$\uparrow$} & {\textbf{R-L}$\uparrow$} & \textbf{MET$\uparrow$} & {\textbf{CIDEr$\uparrow$}}  \\ \specialrule{1.0pt}{0pt}{0.5pt}
        MoEL \cite{moel} & 264.11 & 19.04 & 6.47 & 2.91 & 1.51 & 15.95 & 7.96 & 10.95\\
        MIME \cite{majumder2020mime} & 69.28 & 15.24 & 5.56 & 2.64 & 1.50 & 16.12 & 6.43 & 10.66 \\ 
        EmpDG \cite{li2020empdg} & 115.34  & 18.08 & 6.46 & 3.02 & 1.52 & 15.89 & 6.93 & 10.73\\
        DialoGPT-Joint \cite{liu2021towards} & 15.71 &17.39   & 5.59 & 2.03 & 1.18 & 16.93 &7.55  &11.86 \\
        BlenderBot-Joint \cite{liu2021towards}& 16.79 & 17.62   & 6.91 & 2.81 & 1.66 & 17.94 &7.54  &18.04  \\
        MISC \cite{tu2022misc} & 16.16 & - & 7.31 & - & 2.20 & 17.91 & - & - \\
        GLHG \cite{peng2022control}& 15.67 & 19.66 & 7.57 & 3.74 & 2.13 & 16.37 & - & - \\
        \specialrule{0.7pt}{0.5pt}{0.5pt}
        MultiESC & \textbf{15.41} & \textbf{21.65} & \textbf{9.18} & \textbf{4.99} & \textbf{3.09} & \textbf{20.41} & \textbf{8.84} & \textbf{29.98} \\ 
        MultiESC \emph{w/o} emotion & 18.43 & 18.93 & 7.68 & 4.05 & 2.41 & 20.15 & 7.89 & 24.33 \\
        MultiESC \emph{w/o} cause & 15.68 & 20.07 & 8.76 & 4.64 & 2.77 & 19.82  &8.60  &26.73 \\ 
        MultiESC \emph{w/o} strategy & 15.60 & 19.72 & 8.24 & 4.42 & 2.70 & 20.35 & 8.25 & 27.77 \\
        MultiESC \emph{w/o} lookahead & 15.71 & 21.52 & 9.15 & 4.81 & 3.02 & 20.39 & 8.43 & 29.81 \\
        \specialrule{1.0pt}{0.5pt}{0pt}
		\end{tabular}}
	\end{center}
	\vspace{-2mm}
	\caption{Automatic evaluation results on the generation quality. }
\label{tbl:generation_auto}
\end{table*}

\paragraph{User Feedback Predictor.} \label{sec:UFP}
UFP predicts the user feedback score $f(s_{\ge t}, \mathbf{U}_t)$ by first encoding $s_{\ge t}$ with a Transformer encoder, denoted as $\text{TRS}_{\text{UFP}}$.
Specifically, we leverage a trainable strategy matrix $\mathbf{E}_s \in \mathbb{R}^{|\mathcal{S}| \times d_{\text{emb}}}$ to represent different types of strategies. Given the strategies in $s_{\ge t}$, we concatenate their corresponding strategy vectors as the input of $\text{TRS}_{\text{UFP}}$, so we have 
\begin{equation}
    \mathbf{B} = \text{TRS}_{\text{UFP}}[\text{Emb}(\texttt{[CLS]};s_{\ge t})],\nonumber
\end{equation}
where Emb($\cdot$) represents the operation of the embedding layer that maps the strategies in $s_{\ge t}$ to their corresponding vectors in $\mathbf{E}_s$. 
Suppose the encoded hidden state corresponding to the $\texttt{[CLS]}$ token is $\mathbf{q}_s$. 
Next, we pass the user state embeddings through a Long-Short Term Memory (LSTM) network \cite{cheng2016long}: 
\begin{equation}
\hat{\mathbf{U}}_t = \text{LSTM} (\mathbf{u}_1, \mathbf{u}_2, ..., \mathbf{u}_{t-1}),\nonumber
\end{equation} 
We then use $\mathbf{q_s}$ to attend to the hidden states $\hat{\mathbf{U}}_t$=[$\hat{\mathbf{u}}_1$, $\hat{\mathbf{u}}_2$, ..., $\hat{\mathbf{u}}_{t-1}$] through an attention layer: 
\begin{align}
    \widetilde{\mathbf{u}}_f &= \sum_{i=1}^{t-1} a_i\hat{\mathbf{u}}_i, \nonumber\\
    a_i &= \frac{\exp(\hat{\mathbf{u}}_i^\top\mathbf{W}_a\mathbf{q}_s)}{\sum_{j=1}^{t-1}\exp(\hat{\mathbf{u}}_j^\top\mathbf{W}_a\mathbf{q}_s)},\nonumber
\end{align}
where $\mathbf{W}_a$ is a trainable matrix. 
Finally, we obtain the predicted user feedback score by passing $\widetilde{\mathbf{u}}_f$ through a single feedforward layer.



We leverage the ground-truth user feedback scores annotated in the \textsc{ESConv} dataset as supervision to train the UFP, and use the Mean Squared Error (MSE) as its loss function $\mathcal{L}_{f}$. 

\subsection{Utterance Decoder} 
Given the user state embeddings $\mathbf{U}_{t}$, the dialogue history embeddings $\mathbf{H}_t$, and the selected strategy $\mathbf{\hat{s}}_{t}$, the utterance decoder aims to produce the next utterance $x_{t}$. 
Its architecture is the same as that of the strategy sequence generator (\Cref{sec:SSG}), except for the input sequence. 
To guide dialogue generation with the selected strategy $\mathbf{\hat{s}}_{t}$, we prepend the strategy embedding of $\mathbf{\hat{s}}_{t}$ before the embeddings of the utterance sequence as the input of the utterance decoder. The negative likelihood of the ground-truth token in the target utterance is used as the generation loss $\mathcal{L}_{g}$. More details on the training procedure are provided in the appendix. 


\begin{table*}[t]
    \centering
    \small
    \scalebox{0.96}{
    \begin{tabular}{l | lcr | lcc | lcc |lcc}
        \specialrule{1pt}{0pt}{0.5pt}
        \multirow{2}[0]{*}{\textbf{MultiESC vs.}} & \multicolumn{3}{c|}{\textbf{MoEL}} & \multicolumn{3}{c|}{\textbf{BlenderBot-Joint}} & \multicolumn{3}{c|}{\textbf{w/o strategy}}  & \multicolumn{3}{c}{\textbf{w/o lookahead}} \\
              & {Win} & {Lose} & {Tie$\;$} & {Win} & {Lose} & {Tie} & {Win} & {Lose}  & {Tie}& {Win} & {Lose}  & {Tie}\\
        \specialrule{1.0pt}{0.5pt}{0.5pt}
         {Fluency} & \textbf{64.1}$^{\ddag}$ & 18.0   & 18.0  & 35.2 & \textbf{42.9} & 21.9  & 38.3 & \textbf{41.4} & 20.3 &  \textbf{41.4}  & 37.5 & 21.1 \\
        {Empathy} &  \textbf{53.1}$^{\ddag}$  & 34.4  & 12.5 & \textbf{44.5} & 43.8   & 11.7 & \textbf{43.8}$^{\dag}$ & 29.7  & 26.5  &  35.9  & \textbf{39.1} & 25.0\\
        {Identification} & \textbf{69.5}$^{\ddag}$  & 22.7   & 7.9 &  \textbf{48.4}$^{\ddag}$  & 32.8 & 18.8 &  \textbf{56.3}$^{\ddag}$  & 32.8 & 10.9  & \textbf{46.9}$^{\ddag}$  & 28.9 & 14.2\\
        {Suggestion} &  \textbf{71.9}$^{\ddag}$  & 14.8   & 13.3 &  \textbf{60.9}$^{\ddag}$  & 23.4  & 15.6 &  \textbf{52.3}$^{\dag}$  & 36.7  & 10.9 & \textbf{44.5}$^{\dag}$  & 30.5 & 25.0 \\
        \specialrule{0.7pt}{0.5pt}{0.5pt}
        {Overall} &  \textbf{65.6}$^{\ddag}$  & 20.3  & 14.1 &  \textbf{58.6}$^{\ddag}$  & 31.3 & 10.2 &  \textbf{55.5}$^{\ddag}$  & 30.5 & 14.0 & \textbf{46.1}$^{\dag}$  & 32.0 & 21.9 \\
        \specialrule{1pt}{0.5pt}{0pt}
    \end{tabular}%
    }
    \vspace{-1mm}
    \caption{Human interactive evaluation results (\%). The columns of ``Win/Lose'' indicate the proportion of cases where MultiESC wins/loses in the comparison. ${\dag} / {\ddag}$ denote $p$-value $< 0.1 / 0.05$ (statistical significance test).}
    \label{tbl:generation_human}
\end{table*}

\section{Experiments}
\subsection{Experimental Setup}\label{sec:exp_setup}

\paragraph{Baselines.} 
Our baselines include three empathetic response generators: 
\textbf{MoEL} \cite{moel}, \textbf{MIME} \cite{majumder2020mime}, and \textbf{EmpDG} \cite{li2020empdg}; and four state-of-the-art methods on the \textsc{ESConv} dataset: \textbf{DialoGPT-Joint}, \textbf{BlenderBot-Joint} \cite{liu2021towards}, \textbf{MISC} \cite{tu2022misc}, and \textbf{GLHG} \cite{peng2022control}. More details about them are described in the appendix.

\paragraph{Implementation Details.}
We follow the original division of \textsc{ESConv} for training, validation, and testing. 
We initialize the parameters of the dialogue encoder and the utterance decoder of MultiESC with the BART-small \cite{lewis2020bart} model from the HuggingFace library \cite{Wolf2019huggingface}. 
There are $n_{\text{emo}}$=$65$ types of emotion vectors, with $n_\text{V}$=$n_\text{A}$=8. 
In the strategy planning module, we set $\lambda$=0.7 and $L$=2. 
The beam size $k$ is set to be 6 when searching the set of the most possible strategy sequences $\hat{\mathbf{S}}_L$. 
Since the codes of MISC and GLHG were not released, we directly refer to the results reported in their original papers. For the other baselines, we use their released codes to conduct our experiments. Our model has 145.6M parameters, which is in the same order as the baselines. For reference, BlenderBot-Joint, DialoGPT-Joint, and GLHG have 90M, 117M, and 92M parameters, respectively. More implementation details are provided in the appendix. 

\subsection{Automatic Evaluation of Generation Quality} \label{sec:eval_generation}
To evaluate the generation quality, we adopt the following metrics: perplexity (\textbf{PPL}), BLEU-1/2/3/4 (\textbf{B-1/2/3/4}) \cite{papineni2002bleu}, ROUGE-L (\textbf{R-L}) \cite{lin2004rouge}, METEOR (\textbf{MET}) \cite{Lavie2007Meteor}, and \textbf{CIDEr} \cite{vedantam2015cider}. 
\paragraph{Comparison with Baselines.} 
As shown in the upper part of Table \ref{tbl:generation_auto}, MultiESC performs significantly better than the baseline models in all the metrics. It performs exceptionally well in the CIDEr metric, which measures the similarity between TF-IDF weighted n-grams (i.e., the words that frequently appear in many utterances contribute less to the score). This result demonstrates that MultiESC is more capable of including critical information in the responses, catering to particular situations of users. Another finding is that the three empathetic generators (i.e., MoEL, MIME, and EmpDG) achieve significantly worse perplexity and CIDEr scores than the other models. Through analysis, we find that they tend to include content that commonly appears in many samples (e.g., “I’m sorry to hear that”, “I can understand that”). This is probably because they only focus on how to generate responses that can display an understanding of the user’s emotion, which is insufficient for ESC.

\begin{table}[t]
\small
	\begin{center}\scalebox{0.95}{
		\begin{tabular}{ l  c  c  c } \specialrule{1.0pt}{1pt}{1pt}
        {\textbf{Model}} & {\textbf{Accuracy}} &  {\textbf{Weighted-F1}} & {\textbf{Feedback}} \\ \specialrule{1.0pt}{0pt}{0.5pt} 
         DialoGPT-Joint &  26.03 & 23.86 & 2.87\\
         BlenderBot-Joint & 29.92 & 29.56 & 3.05\\
         MISC & 31.61 & - & -\\
         \specialrule{0.7pt}{0.5pt}{0.5pt} 
        MultiESC  & \textbf{42.01} & \textbf{34.01} & {3.85}\\
         \specialrule{1.0pt}{0.5pt}{0pt}
		\end{tabular}}
	\end{center}\vspace{-2mm}
	\caption{The strategy planning performance of MultiESC and the baseline methods.}
\label{tbl:strategy_acc_baseline}
\end{table}

\paragraph{Ablation Study.}
To analyze the effects of different components on the downstream generation, we compare {MultiESC} with its following variants: (1) \textbf{\emph{w/o} emotion} does not incorporate the emotion embedding layer in the user state modeling module; (2) \textbf{\emph{w/o} cause} does not incorporate emotion cause extraction for user state modeling; (3) \textbf{\emph{w/o} strategy} directly generates utterances without first predicting the used strategy; (4) \textbf{\emph{w/o} lookahead} conducts strategy planning without the lookahead heuristics to estimate the future user feedback scores. 

As shown in the lower part of Table \ref{tbl:generation_auto}, the ablation of any component can cause a drop in the automatic evaluation results, demonstrating the indispensability of each part. In comparison, the ablation of the emotion embedding layer (``\emph{w/o} emotion'') leads to the most significant performance drop, as understanding the user’s emotional states plays the most central role in ESC. The difference between the automatic evaluation results of the full model and ``\emph{w/o} lookahead'' is relatively small. It is because they would generate exactly the same responses if they select the same strategy. The effects of lookahead planning heuristics are more evident in the human interactive test, since the adoption of different strategies at one turn would trigger different responses from the user and lead to different dialogue directions in the future rounds.

\begin{table}[t]
\small
	\begin{center}\scalebox{0.95}{
		\begin{tabular}{ l  c  c  c } \specialrule{1.0pt}{1pt}{1pt}
        {\textbf{Model}} & {\textbf{Accuracy}} &  {\textbf{Weighted-F1}} & {\textbf{Feedback}} \\ \specialrule{1.0pt}{0pt}{0.5pt} 
        MultiESC$_{\text{}k= 1}$ & 38.72  & 30.12 & {3.59}\\
        MultiESC$_{\text{}k= 2}$  & {39.53} & {30.61} & {3.62}\\
        MultiESC$_{\text{}k= 3}$  & {41.33} & {32.83} & {3.75}\\
        MultiESC$_{\text{}k= 4}$  & {41.61} & {33.30} & {3.67}\\
        MultiESC$_{\text{}k= 5}$  & {41.78} & {33.64} & \textbf{3.93}\\
        MultiESC$_{\text{}k= 7}$  & {41.79} & {33.92} & {3.88}\\
        MultiESC$_{\text{}k= 8}$  & {41.79} & {33.97} & {3.92}\\
        \specialrule{0.7pt}{0.5pt}{0.5pt} 
        MultiESC  & \textbf{42.01} & \textbf{34.01} & {3.85}\\
        
        \emph{w/o} lookahead & 38.76 & 30.21 & 3.36\\
         \specialrule{1.0pt}{0.5pt}{0pt}
		\end{tabular}}
	\end{center}\vspace{-2mm}
	\caption{The strategy planning performance of different variants of MultiESC. Note that the beam size of MultiESC is set to be 6 (see \Cref{sec:exp_setup}). }
\label{tbl:strategy_acc_MultiESC}
\end{table}

\subsection{Human Interactive Evaluation}  
We recruit four graduate students with linguistic or psychological background as annotators to chat with the models for human interactive evaluation. We randomly sample 128 dialogues from the test set of \textsc{ESConv}. The annotators are asked to act as the support seekers in these dialogue samples by learning their situations and simulating their process of seeking emotional support by chatting with the models. 
Given MultiESC and a compared model, the annotators are asked to choose which one performs better (or select \emph{tie}) in terms of the following metrics, following \citet{liu2021towards}: 
    (1) \textbf{Fluency}: which model generates more fluent and understandable responses; 
    (2) \textbf{Empathy}: which model has more appropriate emotion reactions, such as warmth, compassion and concern; 
    (3) \textbf{Identification}: which model explores the user's situation more effectively to identify the problem; 
    (4) \textbf{Suggestion}: which model offers more helpful suggestions; 
    (5) \textbf{Overall}: which model provides more effective emotional support overall.

As shown in Table \ref{tbl:generation_human}, we can see that the advantage of MultiESC over MoEL is substantial in all the metrics. 
It also outperforms BlenderBot-Joint in the overall supporting effects, though relatively inferior in terms of fluency, probably because the backbone of BlenderBot-Joint is extensively pre-trained on large-scale dialogue corpora \cite{roller-etal-2021-recipes}. 
Compared with ``\emph{w/o} strategy'', MultiESC is able to show more empathy, more clearly inquire about the user's situation, and provide more specific suggestions, demonstrating the importance of explicit strategy planning in ESC. 
Comparing MultiESC with ``\emph{w/o} lookahead'', we can see that the incorporation of lookahead heuristics brings significant improvement in the dimensions of \emph{identification} and \emph{suggestion}. 

\subsection{Analysis of Strategy Planning}

\begin{figure}[t]
    \centering
    \includegraphics[width=\linewidth]{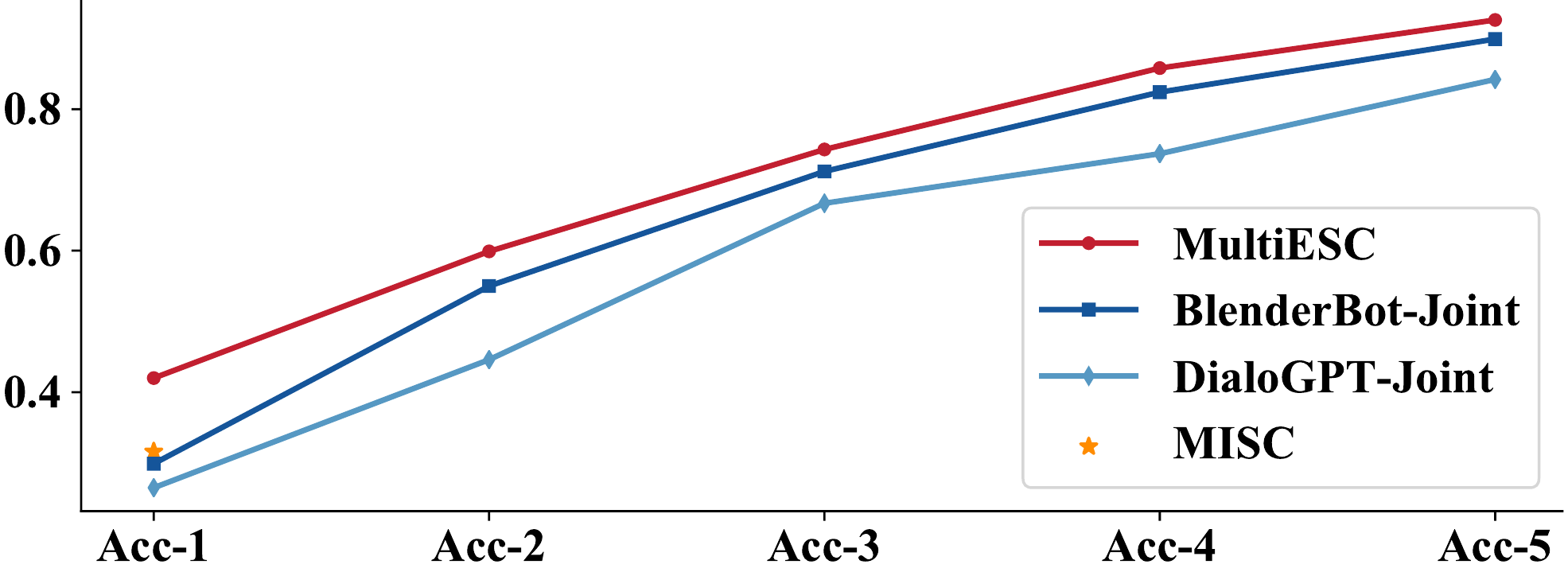}
    \caption{The top-$n$ strategy prediction accuracy of MultiESC and the baseline methods.}
    \label{fig:tok-p acc}
\end{figure}

We evaluate the strategy planning module individually, using the following metrics: \textbf{Accuracy}, the proportion of prediction results that are the same as the ground-truth labels; \textbf{Weighted F1}, the weighted average of F1 scores in different classes while considering the class imbalance; \textbf{Feedback}, the next user feedback score that would be given after the predicted strategy is adopted, simulated with an user feedback predictor as illustrated in \Cref{sec:strategyMultiESC}. 

\paragraph{Comparison with Baselines.} We compare MultiESC with the three baselinses capable of strategy planning (i.e., DialoGPT-Joint, BlenderBot-Joint, and MISC). The results are shown in Table \ref{tbl:strategy_acc_baseline}. We can see that MultiESC performs the best in all the metrics with an absolute improvement of 10.4\% and 4.45\% in accuracy and weighted F1, respectively. 
As shown in Fig. \ref{fig:tok-p acc}, MultiESC also surpasses the baselines in all the top-$n$ accuracy. 

\paragraph{Analysis of MultiESC Variants.} We analyze the following variants of our strategy planning method: (1) \textbf{MultiESC$_{k=?}$}: the model with different beam sizes when searching the set of $k$ most possible strategy sequences $\mathbf{\hat{S}}_L$; (2) \textbf{\emph{w/o} lookahead}: the model without the lookahead heuristics. 
As shown in Table \ref{tbl:strategy_acc_MultiESC}, the strategy planning performance steadily improves with the increase of the beam search size when $k\le$6, as the larger beam size can result in a more precise estimation of the future user feedback. Nevertheless, further increasing $k$ to  consider more strategy sequences of low probabilities does not continue improving the performance apparently when $k$>6. 
Our full model also significantly outperforms ``\emph{w/o} lookahead'' in all the metrics, especially regarding the feedback score. It demonstrates that our lookahead heuristics can help the model better plan the conversation and provide more effective emotional support.

\begin{table}[t]
    \small
    \centering\scalebox{0.95}{
        \begin{tabular}{m{0.97\hsize}}
        \specialrule{1pt}{.5pt}{.5pt}
        \textbf{Bot}: What can I help you with?\\
        \specialrule{0pt}{.5pt}{.5pt}
        
        \textbf{User}: Lately, my friends started to make jokes about my face being ugly. They even called my face as flat as a pan! What should i do? I see them everyday!\\\specialrule{1pt}{1pt}{1pt}
        
        \textbf{MoEL}: I'm sorry to hear that. Have you been able to find that it would be better to let yourself know how you feel about it. \\
        \specialrule{0pt}{1pt}{1pt}
        \textbf{BlenderBot-Joint}: You should tell them to calm down.  \\
        \specialrule{0pt}{1pt}{1pt}
        \textbf{MultiESC}: (\emph{Question}) That's horrible! Have you tried talking to them about it?\\
        \specialrule{0pt}{1pt}{1pt}
        \textbf{MultiESC \emph{w/o} lookahead}: (\emph{Self-disclosure}) I can understand how that would make you feel. I have had to deal with a lot of bullies and I know how hard it can be. \\
        \specialrule{0pt}{1pt}{1pt}
        \textbf{MultiESC \emph{w/o} strategy}: I can understand that. I know why that would be upsetting. \\
        \specialrule{0pt}{1pt}{1pt}
        \textbf{Ground-Truth}: (\emph{Affirmation and Reassurance}) I'm sure you are very handsome! They could just be jealous.\\
        \specialrule{1pt}{.5pt}{.5pt}   
        \end{tabular}}
    \caption{Case study. The upper part is an example of the dialogue history in the test set of \textsc{ESConv}. The lower part shows the responses from different models. }
    \label{tab:case}
\end{table}
\subsection{Case Study}
Table \ref{tab:case} presents a case study of the responses generated by different models. We can see that the utterances from MultiESC and its two variants are more consistent with the context and more empathetic than those of the two baseline models. Further comparing MultiESC and its two variants, the utterance from the ``\emph{w/o} strategy'' seems general and less engaging. The responses generated by MutiESC and ``\emph{w/o} lookahead'' are both of high quality. Nevertheless, with the incorporation of the lookahead heuristic, MultiESC tends to proactively explore the user's situation at the beginning stage of the conversation instead of directly comforting the user, which aligns with the suggested procedure for providing emotional support \cite{hill2009helping}.

\section{Conclusion}
In this paper, we explored the task of developing multi-turn Emotional Support Conversation (ESC) systems, with focus on how to strategically plan the conversation procedure. 
To this end, we proposed a novel ESC system, {MultiESC}, that conducts strategy planning with lookahead heuristics to estimate the long-term effect of the adopted strategy on the user. 
Moreover, we also proposed some effective mechanisms to dynamically model the user's state in multi-turn ESCs. 
The empirical results showed that {MultiESC} achieves significant improvement compared with a set of strong baselines in both generation quality and strategy planning. 

\section*{Limitations}

Though our proposed method exhibits large improvement compared with the existing baselines, we believe that the research on emotional support chatbots still has a long way to go. Compared with the human supporters, the utterances generated by the chatbots are usually general and repetitive, unable to show a personalized, in-depth understanding of the user's experiences or provide very specific and constructive suggestions on how to change the situation. This issue might be alleviated through the incorporation of commonsense knowledge, which will be included in our future research direction. Other issues, such as how to construct more trustworthy and safe emotional support chatbots, also require much further exploration. 

\section*{Ethical Considerations}
It needs to be clarified that the term ``emotional support'' in our paper mainly refers to peer support, like the one we can seek from family and friends in daily conversation.  We do not claim to construct chatbots that can provide professional psycho-counseling or psychological treatment. Still, it needs particular caution when using such systems, and considerable further efforts are required to construct safer ESC systems. For example, crisis-warning mechanisms to detect users who have tendencies of self-harming or suicide are desirable. 

Our experimental dataset, ESConv, is a well-established, publicly-available benchmark. It has filtered the sensitive and private information during the dataset construction. 
The participants in our human evaluation were transparently informed of our research intent and were paid reasonable wages. 

\section*{Acknowledgments}
This work was supported by the Research Grants Council of Hong Kong (PolyU/5204018, PolyU/15207920, PolyU/15207122) and National Natural Science Foundation of China (62076212). 
It was also supported in part by in part by National Key R\&D Program of China under Grant No. 2020AAA0109700. 
We also thank Jian Wang and Yueyuan Li for their helpful comments.

\bibliographystyle{acl_natbib}
\bibliography{acl2021}

\newpage
\section*{Appendix}
\subsection*{A. Training Procedure}

For training of the strategy sequence generator, we use the negative log-likelihood of the ground-truth strategy $s_{_{t+l}}^{*}$ as its loss function $\mathcal{L}_{s}$. For the utterance decoder, the negative likelihood of the ground-truth token in the target utterance is used as the generation loss $\mathcal{L}_{g}$. The strategy sequence generator and the utterance decoder are trained jointly, with the total loss as $\mathcal{L} = \mathcal{L}_{s} + \mathcal{L}_{g}$. 
For training of the user feedback predictor, we use the Mean Squared Error (MSE) as its loss function $\mathcal{L}_{f}$. Since the user feedback scores in the \textsc{ESConv} dataset are mainly between 2-5, we augment the training data by randomly generating 5,000 strategy sequences and regarding them as samples with the score 1. It is trained independently from the strategy predictor and the utterance decoder. 

\subsection*{B. Baselines}
\noindent \textbf{MoEL} \cite{moel} adopts several decoders focusing on different types of emotional utterances, whose outputs are combined to generate the final utterances. 

\noindent \textbf{MIME} \cite{majumder2020mime} follows the architecture of MoEL and adds extra mechanisms to combine the results from different decoders. 

\noindent \textbf{EmpDG} \cite{li2020empdg} learns how to generate responses consistent with the user's emotion via an adversarial learning framework. 

\noindent \textbf{DialoGPT-Joint} and \textbf{BlenderBot-Joint} \cite{liu2021towards} are developed on the backbones of DialoGPT \cite{zhang2020dialogpt} and BlenderBot \cite{roller-etal-2021-recipes}, respectively. They prepend a special token, denoting the predicted support strategy, before the generated utterance to generate content conditioned on a predicted strategy. 

\noindent \textbf{MISC} \cite{tu2022misc} enhances context encoding with commonsense knowledge and uses the predicted strategy distribution to guide the emotional support dialogue generation. It predicts the strategy distribution using a vanilla Transformer encoder. 

\noindent \textbf{GLHG} \cite{peng2022control}  adopts a graph neural network to model the relationships between the user's emotion causes, intentions and the dialogue history for emotional support dialogue generation.

\subsection*{C. Implementation Details}
We initialize the parameters of the dialogue encoder and the utterance decoder of MultiESC with the BART-small \cite{lewis2020bart} model from the HuggingFace library \cite{Wolf2019huggingface}. 
The maximum length of the input sequence for the dialogue encoder is $N$=$512$. The dimensions of all the hidden embeddings are $d_{\text{emb}}$=$768$. 
There are $n_{\text{emo}}$=$65$ types of emotion vectors, with $n_\text{V}$=$n_\text{A}$=8. 

In the strategy planning module, we set $\lambda$=0.7, which results in the best performance on the validation set among $\lambda\in$\{0.1, 0.2, ..., 1.0\}. 
For the number of lookahead rounds $L$, we experiment with $L\in$\{1, 2 ,.., 5\}. We find that the performance on the validation set is the best when $L=$2 and $L=$3. The performances when $L=$2 and $L=$3 are very close, but considering the computation efficiency, we set $L=$2 in the following experiments. 
For the searching beam size $k$, we experiment with $k\in$\{1, 2 ,.., 10\}, and finally set it to be 6, because it strikes the best balance between performance and efficiency. We choose the above hyperparameters by manual tuning, and the selection criterion is the strategy prediction accuracy on the validation set. 

AdamW \cite{Losh2017AdamW} is used as optimizer; its initial learning rate is set to be $5$$\times$$10^{-5}$ and adaptively decays during training. 
The batch size is set to be 32.  Since the codes of MISC and GLHG were not released, we directly refer to the results reported in their original papers. For the other baselines, we use their released codes to conduct our experiments.
Each model is trained up to 10 epochs, and the checkpoints that achieve the best perplexity on the validation set are used for evaluation. 

Our model has 145.6M parameters. 
The hardware we used was one GPU of NVIDIA Tesla V100. The overall training time is about two hours.

\subsection*{D. Dataset}

Our experiments are conducted on the \textsc{ESConv} dataset \cite{liu2021towards}\footnote{\url{https://github.com/thu-coai/Emotional-Support-Conversation}}. It is an English dataset. 
To construct the dataset, they recruited crowd-workers, who had learned the common procedures and strategies for providing emotional support, to converse with volunteers that needed emotion support through an online platform. 
The crowd-workers were required to annotate the strategy adopted at each turn, and the support-seekers were asked to give feedback every two rounds on a Likert scale with five levels, indicating how much their emotional distress is reduced. The dataset contains 1,300 long dialogues with 38,350 utterances.  There is an average of 29.5 utterances per dialogue and an average 16.7 tokens per utterance. We follow the original division of the \textsc{ESConv} dataset for training, validation, and testing, with the ration of 8:1:1. There are overall 8 types of support strategies. Referring to \citet{liu2021towards}, their original definitions are as follows:

\begin{table}[t]
\small
    \centering
      \scalebox{1.0}{
        \begin{tabular}{lr}
        \specialrule{1.0pt}{.5pt}{0pt}    
        \textbf{Strategy} & \textbf{Proportion}\\\specialrule{1.0pt}{.5pt}{0pt}    
         Question & 21.77\% \\
          Restatement or Paraphrasing & 6.46\% \\
          Reflection of Feelings & 8.05\% \\
          Self-disclosure  & 9.34\% \\
          Affirmation and Reassurance  & 16.13\% \\
          Providing Suggestions or Information& 22.02\% \\
          Greetings & 8.72\% \\
          Others (Unlabelled) & 7.49\% \\
        \specialrule{1.0pt}{.5pt}{0pt}    
        \end{tabular}%
      }
    \caption{The strategy distribution.}
    \label{tab:strategy}
\end{table}

\begin{itemize}
    \item \emph{Question}: ask for information related to the problem to help the help-seeker articulate the issues that they face.  
    \item \emph{Restatement or Paraphrasing}: 
    a simple, more concise rephrasing of the support-seeker’s statements that could help them see their situation more clearly.
    \item \emph{Reflection of Feelings}: describe the help-seeker’s feelings to show the understanding and empathy.
    \item \emph{Self-disclosure}: share similar experiences or emotions that the supporter has also experienced to express your empathy.
    \item \emph{Affirmation and Reassurance}: affirm the support-seeker’s ideas, capabilities, and strengths to give reassurance and encouragement. 
    \item \emph{Providing Suggestions}: provide suggestions about how to change the current situation.
    \item \emph{Information}: provide useful information to the help-seeker, for example with data, facts, opinions, resources, or by answering questions.
    \item \emph{Others}: other support strategies that do not fall into the above categories.
\end{itemize}

In our experiments, we make some adaptions to the strategy annotation. On one hand, we find that the definition of \emph{Providing Suggestions} and \emph{Information} are often hard to differentiate. As shown in Table \ref{tbl:example}, some responses annotated as \emph{Providing Suggestions} can also be regarded as providing useful information, and those labelled as \emph{Information} also offer suggestions. Thus, we merge these two categories into one type of strategy, named as ``\emph{Providing Suggestions or Information}''. 

On the other hand, we find that the existence of the ``\emph{Others}'' category largely impedes the model performance. As illustrated in the above definition, some responses in this category are exchange of pleasantries, which is the case for approximately 50\% of the responses annotated as \emph{Others}. We argue that the exchange of pleasantries is also an essential strategy, as it can help to establish a friendly connection with the user \cite{miller2012motivational}. Thus, we define a new strategy category, named as ``\emph{Greetings}'' for such kind of responses. We obtain the annotation of them by first using a set of regular expression matches and then manually double-checking the results. For the rest of responses labelled as \emph{Others}, we directly regard them as unlabelled data and do not include them in the training of the strategy planning module, because we find that many of them can actually be classified as the other types of strategies but were mislabelled, and this strategy are not helpful in enhancing the response generation. 
The strategy distribution after the above adaptions is presented in Table \ref{tab:strategy}.

\begin{table}[t]
\footnotesize
	\begin{center}
        \begin{tabular}{m{3.4cm}m{3.4cm}}\specialrule{1.0pt}{0pt}{1pt}
\multicolumn{1}{c}{\textbf{\emph{Providing Suggestions}}}   & \multicolumn{1}{c}{\textbf{\emph{Information}}}                        \\ \specialrule{1.0pt}{.5pt}{.5pt}
I understand that. What if that is the case? You may need to talk to them and let them know how you feel about that. &
Is it possible to reframe how you look at your clients' dire financial situations? 
 \\ \specialrule{0.7pt}{.5pt}{.5pt}
Have you thought of contacting a debt relief program? In some cases they can substantially reduce debt to something much more manageable.  & It may not be for you. I think you should think about the pros and cons of keeping your position. It might make things clearer for you.\\
        \specialrule{1.0pt}{.5pt}{0pt}        
        \end{tabular}
	\end{center}
\caption{Examples of the responses labelled with the strategies \emph{Providing Suggestions} and \emph{Information} in the original ESConv dataset.}
\label{tbl:example}
\end{table}
\end{document}